\pdfoutput=1

\documentclass[11pt]{article}

\usepackage{ACL2023}

\usepackage{times}
\usepackage{latexsym}
\usepackage{graphicx}
\usepackage{subcaption}
\usepackage{booktabs}
\usepackage{multirow}
\usepackage{array}
\usepackage{xcolor}

\usepackage[T1]{fontenc}

\usepackage[utf8]{inputenc}

\usepackage{microtype}

\usepackage{inconsolata}

\usepackage{amsmath, amssymb, amsfonts, mathtools, bm}

\title{Instructions for ACL 2023 Proceedings}

\author{
    Shruti Singh Baghel\textsuperscript{1} \and
    Yash Pratap Singh Rathore\textsuperscript{1} \and Sushovan Jena\textsuperscript{1} \and Anurag Pradhan\textsuperscript{2}
     \AND
    \and
    Amit Shukla\textsuperscript{1} \and
    Arnav Bhavsar\textsuperscript{1} \and
    Pawan Goyal\textsuperscript{4} \\
    \\
    \textsuperscript{1}Indian Institute of Technology Mandi, \texttt{\{s24110,s24036, s20011, amitshukla, arnav\}@iitmandi.ac.in} \\
    \textsuperscript{2}Vellore Institute of Technology, \texttt{anurag.pradhan2023@vitstudent.ac.in} \\
    \textsuperscript{4}Indian Institute of Technology Kharagpur, \texttt{pawang@cse.iitkgp.ac.in}
}

\title{Towards Blind and Low-Vision Accessibility of Lightweight VLMs and Custom LLM-Evals}

\begin{document}
\maketitle

\begin{abstract}
Large Vision-Language Models (VLMs) excel at understanding and generating video descriptions but their high memory, computation, and deployment demands hinder practical use particularly for blind and low-vision (BLV) users who depend on detailed, context-aware descriptions. To study the
effect of model size on accessibility-focused description quality, we
evaluate SmolVLM2 variants with 500M and 2.2B parameters across
two diverse datasets: AVCaps (outdoor), and Charades (indoor).
In this work, we introduce two novel evaluation frameworks
specifically designed for BLV accessibility assessment: the Multi-Context BLV Framework evaluating spatial orientation, social interaction, action events, and ambience contexts; and the Navigational
Assistance Framework focusing on mobility-critical information.
Additionally, we conduct a systematic evaluation of four different
prompt design strategies and deploy both models on a smartphone,
evaluating FP32 and INT8 precision variants to assess real-world
performance constraints on resource-limited mobile devices.
\end{abstract}

\section{Introduction}
Large multimodal vision-language models (VLMs), such as GPT and LLaVA series by OpenAI and Microsoft\cite{liu2023llava,openai2023gpt4}, have shown impressive capabilities in understanding and generating detailed descriptions of visual content. 
While these large models can produce high quality audio descriptions that align with professional standards, their practical application is restricted by their high computational requirements, dependence on cloud infrastructure, which requires high internet bandwidth 
making them unsuitable for deployment on everyday devices such as mobile phones or tablets. This renders them impractical for BLV users from experiencing real time, private, on-device accessibility. 

Our research investigates if significantly smaller models, which can operate on resource-limited devices, can generate video descriptions that are comparable in quality to those produced by large, resource-heavy ones.
In real world settings, BLV users require on device solutions capable of providing timely and detailed descriptions without relying on remote servers or continuous internet connectivity. A lightweight model integrated into a smartphone application could locally process live or pre-recorded video, enabling synchronized and context-aware audio feedback such as scene changes, object appearances, and actions delivered directly through headphones. 

Small vision-language models are emerging as a promising approach to overcome the drawbacks of larger models while still delivering competitive performance on specific tasks. These compact models, usually with fewer than 2 billion parameters, capable of operating effectively on consumer-grade hardware, enabling on-device implementation and real-time processing. SmolVLM2-500M-Video-Instruct \cite{allal2024smolvlm} and SmolVLM2-2.2B-Video-Instruct\cite{marafioti2025smolvlm} are notable developments in this area, tailored for video understanding tasks. 

Furthermore, human annotations (HA) and contextual information are integrated to enhance model understanding providing comprehensive guidance for accessibility-focused video description generation. However, they frequently fall short for BLV users who need precise, contextually relevant, and in-depth information. To address these limitations, professional audio-description (AD) guidelines developed by organizations such as Netflix, Ofcom, Media Access Canada, and the Described and Captioned Media Program\cite{li2025videoa11y} provide structured frameworks that ensure consistency in character identification, scene description, and narrative flow comprehension. As illustrated in Figure 1, SmolVLM2-500M-Video-Instruct generates increasingly detailed and accessibility-focused descriptions when enhanced with human annotations (HA) and professional AD guidelines.

To validate practical deployment viability, we conducted real-world testing on a mobile device, evaluating both SmolVLM2 variants in FP32 and INT8 precision formats.This on-device deployment approach demonstrates that professional-quality video descriptions can be generated locally on consumer devices without cloud connectivity, establishing feasibility for democratizing video accessibility for BLV users.\newline

\textbf{Key Contributions:}

(1){We evaluated SmolVLM2 variants across two different environmental contexts, revealing smaller models often outperform larger variants in specific accessibility scenarios.}

(2){We implement four progressive prompting strategies to investigate how instruction complexity affects model performance for BLV users.} 

(3){We introduce two specialized evaluation frameworks, the Multi-Context Evaluation Framework and Navigation Assistance Framework that address critical gaps in existing evaluation methodologies which currently undervalue BLV users' preferences.}

(4){We demonstrate that professional quality audio-descriptions may be produced locally without relying on the cloud through extensive real-world deployment testing on consumer-grade smartphones.}


\section{DATASETS}

Our evaluation utilizes two benchmark datasets that represent two different environmental contexts(indoor and outdoor). 
\begin{itemize}
\item{Indoor Dataset} \cite{sigurdsson2016charades}: From the original 9,848 videos (7,985 training, 1,863 testing), we selected 498 videos and their corresponding human annotations from the test set. This represents approximately 27\% of the test set, chosen to include diverse indoor activities while ensuring balanced representation across activity categories (cooking, cleaning activities, etc.).
\item {Outdoor Dataset} \cite{sudarsanam2024avcaps}: We selected 423 outdoor videos and their human annotations from the complete collection of 2,061 clips across all partitions. This 20\% sample was stratified across different outdoor scenarios (urban environments, parks, streets, natural settings) to maintain environmental diversity crucial for evaluating outdoor navigation assistance.

We selected this smaller subset of the dataset to evaluate the model’s performance in diverse real-world scenarios with varying lighting, weather, and background complexity.
\end{itemize}

\section{Framework}
Our research investigates the performance trade-offs between resource-constrained and resource-intensive vision-language models for accessibility -focused video description. We design a comprehensive evaluation framework that systematically compares SmolVLM2-500M-Video-Instruct \cite{allal2024smolvlm} and SmolVLM2-2.2B \cite{marafioti2025smolvlm} across diverse video content and prompting strategies.

\subsection{Overview}
Our approach enables systematic investigation of how model size affects accessibility-focused video description quality across varying instruction complexity levels. Our experimental design employs four distinct prompting strategies that demonstrate progressive complexity from baseline approaches to comprehensive accessibility-focused instruction integration.

\begin{figure*}[!htbp]
    \centering
    \includegraphics[width=0.9\textwidth]{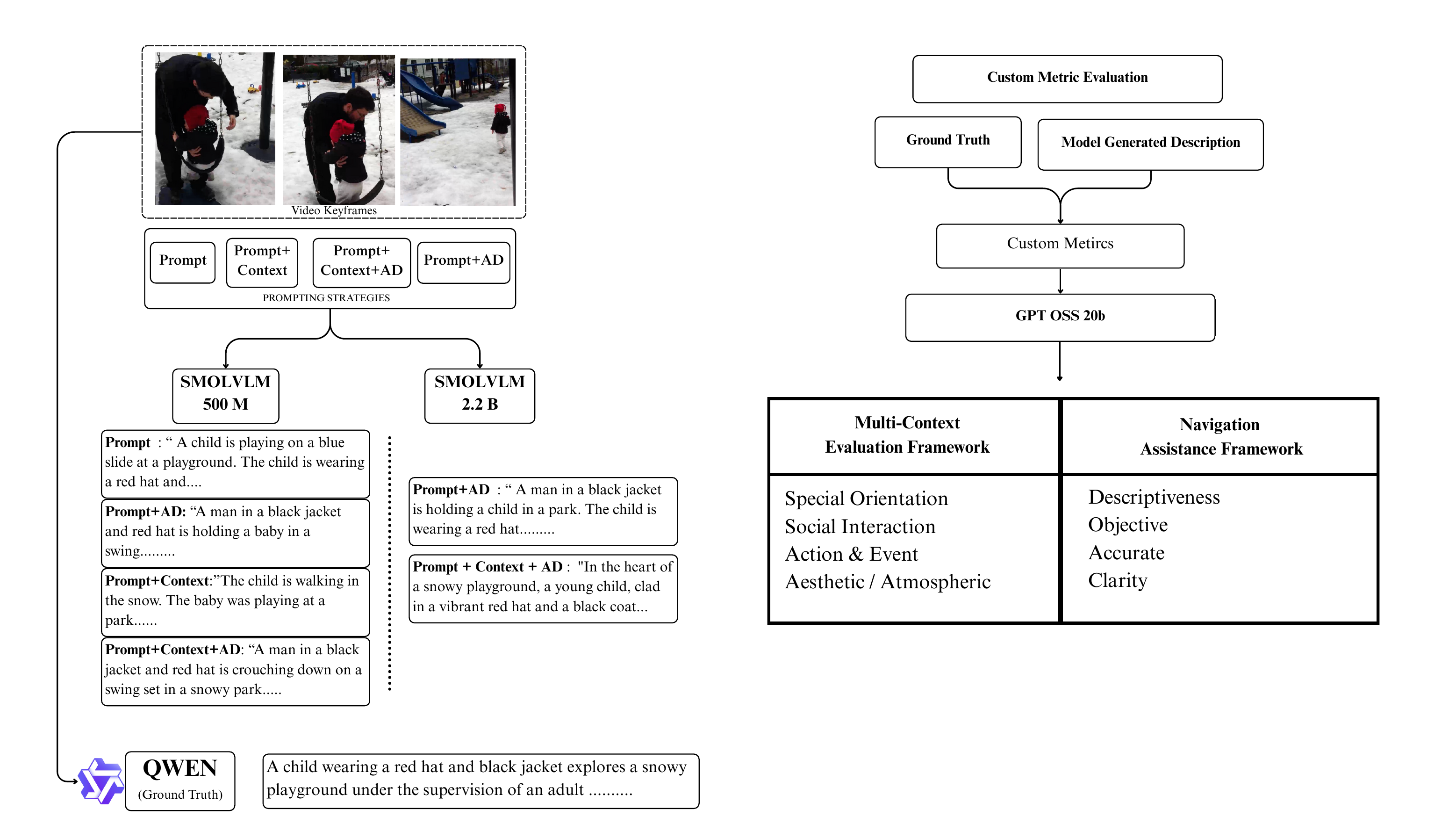}
    \caption{Experimental Design Overview: Four prompting strategies evaluated across SmolVLM variants and reference model (Qwen). The diagram illustrates progressive complexity from baseline prompt-only approach to comprehensive prompt with context and audio-description guidelines integration. Each strategy generates descriptions that are evaluated against ground truth using both standard NLP metrics and custom accessibility metrics designed for BLV users.}
    \label{fig:system_overview}
\end{figure*} 
\subsection{Model Selection}
For our evaluation, we selected SmolVLM2-500M-Video-Instruct and SmolVLM2-2.2B-Video-Instruct due to their combined advantages for video description tasks \cite{marafioti2025smolvlm}. Both models are explicitly fine-tuned for video understanding with temporal mechanisms essential for coherent description generation, while maintaining exceptional edge deployment viability with GPU memory requirements of only 1.8 GB and 5.2 GB respectively-significantly lower than larger alternatives. The 500M variant achieves competitive performance on Video-MME (42.2) with maximum computational efficiency, while the 2.2B variant offers enhanced quality for scenarios with additional resources, both demonstrating state-of-the-art performance in their respective parameter classes. Critically, both variants support robust instruction following capabilities necessary for implementing professional audio-description guidelines from VideoA11y \cite{li2025videoa11y}, enabling real-time inference on consumer hardware and democratizing video accessibility across different computational constraints. 
Specifically, we address three core research questions:
\begin{enumerate}
\item[\textbf{Q1.}] How effectively can small models match large model performance for accessibility-focused video description when guided by professional audio-description (AD) guidelines?
\item[\textbf{Q2.}] How do performance trade-offs affect deployment on resource limited hardware such as smartphones?
\item[\textbf{Q3.}] Why custom accessibility metrics are better than standard NLP metrics in capturing the true preferences of blind and low-vision users for the quality of video descriptions?
\end{enumerate}
This comprehensive evaluation enables us to understand the practical implications of deploying compact VLMs for accessibility applications while maintaining the quality standards necessary for BLV users.

\subsection{Comprehensive Approach}
We use Qwen 2.5 VL 7B Instruct with expert audio-description guidelines from VideoA11y \cite{li2025videoa11y} to generate ground truth which correctly processes all 42 audio-description guidelines and produces descriptions that meet professional accessibility standards, exhibiting strong instruction following capabilities necessary for putting VideoA11y's methodology into practice. 

To efficiently process video content while maintaining essential visual information, we implemented an adaptive keyframe extraction algorithm that analyzes inter-frame differences in the LUV color space. The method computes absolute differences between consecutive frames, applies Hanning window smoothing, and identifies local maxima in the difference signal. 
 In our implementation, we extracted 3-4 keyframes per video, while increasing keyframe density can enhance temporal coverage, it also introduces additional computational cost, a key consideration for on-device deployment.

Following the experimental paradigm established VideoA11y framework \cite{li2025videoa11y}, we use four different prompting techniques to assess how contextual information and instruction complexity affect model performance: (1) {Prompt Only} - utilizing zero-shot generation with the standardized compliant prompt to establish baseline performance without additional guidance. 
(2) {Prompt with Context} - incorporating the compliant prompt with original human annotations from the datasets 
to evaluate the model's ability to leverage existing annotation information. Context refers to human annotations from original datasets (Indoor and Outdoor), exactly as implemented in VideoA11y. Human annotations are concatenated with the prompt as "Current Description" before being fed to the MLLM. 
(3) {Prompt with Context and AD Guidelines} - combining the prompt with human annotations and 42 professional audio-description guidelines to assess comprehensive multimodal instruction following.
(4) {Prompt with AD Guidelines} - integrating the compliant prompt with audio-description guidelines only to test whether structured accessibility guidelines alone can enable compact models to produce descriptions meeting BLV users' requirements. 

\subsection{Proposed Evaluation Frameworks}
Our evaluation protocol addresses the critical limitations of reference based metrics for accessibility applications \cite{kapur2024reference}. We employ dual assessment methodologies: standard NLP metrics for comparison with existing research, and two novel accessibility-centric evaluation frameworks. These frameworks are specifically designed to reflect BLV users' actual needs and preferences. This dual evaluation approach overcomes the systematic bias that reference based metrics exhibit against BLV users' preferences, as demonstrated by Kapur and Kreiss \cite{kapur2024reference}. 
While VideoA11y effectively assess general description quality, they lack granularity for diverse BLV contexts and navigational needs. To fill these gaps, we introduce two complementary frameworks: Mult-Context BLV Framework and Navigational Assiatance Framework. 

\subsubsection{Multi-Context BLV Framework}
\label{sec:multicontext_framework}

This framework evaluates descriptions across four critical user scenarios that reflect diverse BLV information needs in real-world settings:

\begin{itemize}
    \item[(i)] {Spatial Orientation} (1-10 scale): Assesses location descriptions, directional cues, relative positioning, and environmental layout information essential for mental mapping.
    \item[(ii)]  {Social Interaction} (1-10 scale): Evaluates person identification, interpersonal dynamics, emotional expressions, and social context crucial for understanding human interactions.
    \item[(iii)] {Action \& Events} (1-10 scale): Measures temporal sequence clarity, activity description completeness, and causal relationships between events.
    \item[(iv)] {Ambience} (1-10 scale): Captures mood, lighting conditions, environmental atmosphere, and sensory details that enhance immersive comprehension.
\end{itemize}
\begin{equation*}
\begin{split}
\text{MCF\_Score} = \frac{1}{4} \big( 
    S_{\text{spatial}} + 
    S_{\text{social}} + 
    S_{\text{action}} \\
    + S_{\text{Ambience}} 
\big)
\end{split}
\end{equation*}

\noindent where:
\begin{itemize}
    \item $S_{\text{spatial}} \in [1,10]$: Spatial Orientation Context score
    \item $S_{\text{social}} \in [1,10]$: Social Interaction Context score
    \item $S_{\text{action}} \in [1,10]$: Action \& Event Context score
    \item $S_{\text{Ambience}} \in [1,10]$: Ambience Context score
\end{itemize}

Our framework weights these dimensions based on navigation critical scenarios rather than general description quality.
\subsubsection{Navigational Assistance Framework}
\label{sec:navigation_framework}
This framework focuses on mobility critical information through four dimensions essential for spatial navigation and safety:

\begin{itemize}
    \item[(i)] {Descriptiveness}: Spatial layout detail, hazard identification, and environmental feature descriptions (obstacles, pathways, boundaries).
    \item[(ii)] {Objectivity}: Factual reporting without assumptions, avoiding subjective interpretations of spatial relationships.
    \item[(iii)] {Accuracy}: Precision in spatial relationships, object positions, and distance estimations critical for navigation decisions.
    \item[(iv)] {Clarity}: Information organization for sequential navigation decision-making, including logical flow and unambiguous directional references.
\end{itemize}
\begin{equation*}
\begin{split}
\text{NAF\_Score} = \frac{1}{4} \Big(
    N_{\text{descriptiveness}} + 
    N_{\text{objectivity}} \\
    + N_{\text{accuracy}} 
    + N_{\text{clarity}}
\Big)
\end{split}
\end{equation*}

\noindent
\\

where:
\begin{itemize}
    \item $N_{\text{descriptiveness}} \in [1,10]$: Descriptiveness metric score
    \item $N_{\text{objectivity}} \in [1,10]$: Objectivity metric score
    \item $N_{\text{accuracy}} \in [1,10]$: Accuracy metric score
    \item $N_{\text{clarity}} \in [1,10]$: Clarity metric score
\end{itemize}

\subsubsection{Implementation and Validation}
\label{sec:framework_validation}
For custom accessibility metrics evaluation, we employ GPT-OSS-20B\cite{openai2025gptoss}, a 20-billion parameter open-source language model, following the VideoA11y evaluation methodology \cite{li2025videoa11y}. We conducted all evaluations with GPT-OSS-20B running locally via the Ollama\cite{ollama2024} server to ensure offline, reproducible results without network dependencies. The model processes both our Qwen 2.5 VL 7B Instruct generated ground truth descriptions and descriptions produced by both SmolVLM variants (500M and 2.2B) using VideoA11y's standardized evaluation template, enabling consistent assessment of the four custom accessibility dimensions. The systematic evaluation enables investigation of our three core research questions outlined earlier (Section 3.2).

\subsection{Mobile Deployment and Performance Evaluation}
To assess real-world deployment viability for accessibility applications, we conducted comprehensive on-device evaluation using a Vivo Y27 smartphone equipped with a MediaTek Helio G85 octa-core processor and Mali-G52 MC2 GPU with 6GB shared system memory. 
Our deployment methodology employed the llama.cpp framework's llama-mtmd-cli tool, requiring model conversion to .gguf format for mobile compatibility. FP32 variants were converted from their original safetensors format using the official convert\_hf\_to\_gguf.py script, while INT8 quantized versions were generated through Hugging Face's "GGUF My Repo" feature to evaluate precision-performance trade-offs essential for resource-constrained deployment.

The mobile execution environment utilized Termux for Android terminal access, enabling local compilation of llama-mtmd-cli and direct model inference without external dependencies. We implemented a keyframe extraction pipeline using FFmpeg within the mobile environment, processing videos into sequential image frames that were combined with textual prompts incorporating professional AD guidelines.

Both FP32 and INT8 versions of the two models were tested under identical conditions. This setup allowed us to collect detailed performance measurements, including latency, memory usage, and operational behavior during inference on a resource‑constrained mobile platform.

\begin{table*}[!htbp]
\centering
\caption{SmolVLM2-500M-Video-Instruct: Standard NLP Metrics Performance}
\label{tab:smolvlm500m_nlp}
\begin{tabular}{@{}lccccccc@{}}
\toprule
\textbf{Strategy / Dataset} & \textbf{BLEU-1} & \textbf{BLEU-4} & \textbf{METEOR} & \textbf{ROUGE-L} & \textbf{SPICE} & \textbf{CIDEr} \\
\midrule
\textbf{Indoor} & & & & & & \\
Prompt Only & 0.191 & 0.046 & 0.145 & 0.254 & 0.1937 & 0.134 \\
Prompt + Context & 0.304 & 0.062 & 0.112 & 0.251 & 0.1526 & 0.1358 \\
Prompt + AD Guidelines & 0.311 & 0.077 & 0.156 & 0.275 & 0.1795 & 0.1721 \\
Prompt + Context + AD Guidelines & 0.287 & 0.070 & 0.153 & 0.268 & 0.1729 & 0.194 \\
\midrule
\textbf{Outdoor} & & & & & & \\
Prompt Only & 0.135 & 0.029 & 0.139 & 0.235 & 0.1872 & 0.116 \\
Prompt + Context & 0.195 & 0.034 & 0.12 & 0.22 & 0.1553 & 0.137 \\
Prompt + AD Guidelines & 0.223 & 0.047 & 0.148 & 0.251 & 0.1943 & 0.207 \\
Prompt + Context + AD Guidelines & 0.273 & 0.055 & 0.162 & 0.247 & 0.171 & 0.131 \\
\midrule
\end{tabular}
\\[0.05cm]
\footnotesize
\textit{Note: All metrics are scored on a 0-1 scale where higher values indicate better performance.}
\end{table*}

\begin{table*}[!htbp]
\centering
\caption{SmolVLM2-2.2B-Instruct: Standard NLP Metrics Performance}
\label{tab:smolvlm2.2b_nlp}
\begin{tabular}{@{}lcccccc@{}}
\toprule
\textbf{Strategy / Dataset} & \textbf{BLEU-1} & \textbf{BLEU-4} & \textbf{METEOR} & \textbf{ROUGE-L} & \textbf{CIDEr} & \textbf{SPICE} \\
\midrule
\textbf{Indoor} & & & & & & \\
Prompt + AD Guidelines & 0.2723 & 0.0619 & 0.1353 & 0.2606 & 0.1930 & 0.1768 \\
Prompt + Context + AD Guidelines & 0.3271 & 0.0798 & 0.1363 & 0.2750 & 0.2258 & 0.1841 \\
\midrule
\textbf{Outdoor} & & & & & & \\
Prompt + AD Guidelines & 0.1850 & 0.0345 & 0.1515 & 0.1946 & 0.0884 & 0.1462 \\
Prompt + Context + AD Guidelines & 0.1878 & 0.0331 & 0.1485 & 0.1913 & 0.0719 & 0.142 \\
\bottomrule
\end{tabular}
\end{table*}

\begin{table*}[!htbp]
\centering
\caption{SmolVLM2-2.2B-Instruct: Custom Accessibility Metrics Performance}
\label{tab:smolvlm2.2b_custom}
\begin{tabular}{@{}lcccc@{}}
\toprule
\textbf{Strategy / Dataset} & \textbf{Descriptive} & \textbf{Objective} & \textbf{Accurate} & \textbf{Clear} \\
\midrule
\textbf{Indoor} & & & & \\
Prompt + AD Guidelines & 2.508 & 3.25 & 1.935 & 3.345 \\
Prompt + Context + AD Guidelines & 2.529 & 3.246 & 1.78 & 3.414 \\
\midrule
\textbf{Outdoor} & & & & \\
Prompt + AD Guidelines & 2.908 & 2.712 & 1.778 & 3.095 \\
Prompt + Context + AD Guidelines & 2.936 & 2.761 & 1.835 & 3.222 \\
\bottomrule
\end{tabular}
\end{table*}

\begin{table*}[!htbp]
\centering
\caption{\centering Model Performance Comparison:Prompt + Context + AD Guidelines Strategy using custom metrics}
\label{tab:smolvlm500m_custom}
\begin{tabular}{@{}lcccc@{}}
\toprule
& \multicolumn{2}{c}{\textbf{SmolVLM2-500M-Video-Instruct}} & \multicolumn{2}{c}{\textbf{SmolVLM2-2.2B-Instruct}} \\ 
\cmidrule(lr){2-3} \cmidrule(lr){4-5}
\textbf{Metric} & \textbf{Outdoor} & \textbf{Indoor} & \textbf{Outdoor} & \textbf{Indoor} \\
\midrule
Descriptive & 3.031 & 2.779 & 2.936 & 2.529 \\
Objective   & 2.747 & 2.793 & 2.761 & 3.246 \\
Accurate    & 1.719 & 1.627 & 1.835 & 1.780 \\
Clarity     & 3.177 & 3.094 & 3.222 & 3.414 \\
\bottomrule
\end{tabular}
\end{table*}

\begin{table*}[!htbp]
\centering
\caption{Performance Comparison of SmolVLM2 Models with FP32 and INT8 Quantization}
\label{tab:model_performance}
\footnotesize
\setlength{\tabcolsep}{4pt}
\renewcommand{\arraystretch}{1.2}
\begin{tabular}{@{}lccccc@{}}
\toprule
\textbf{Model} & \multicolumn{2}{c}{\textbf{SmolVLM2-500M}} & \multicolumn{2}{c}{\textbf{SmolVLM2-2.2B}} \\ 
\cmidrule(lr){2-3} \cmidrule(lr){4-5}
 & \textbf{FP32} & \textbf{INT8} & \textbf{FP32} & \textbf{INT8} \\
\midrule
\textbf{LATENCY} & 33639.04 & 29904.29 & 2000642.04 & 201306.71 \\
\midrule
\textbf{PEAK DRAM USAGE} & 1142.784 MB & 761.856 MB & 2797.216 MB & 2512.896 MB \\
\midrule
\textbf{MODEL SIZE} & 190.22 MB & 103.73 MB & 831.87 MB & 565.05 MB \\
\midrule
\textbf{TOKEN PER SECOND} & 6.41 & 13.55 & 0.05 & 1.47 \\
\scriptsize (generation speed) & & & & \\
\midrule
\textbf{TIME TO FIRST TOKEN} & 17120.57 ms & 18797.63 ms & 150457.48 ms & 123936.97 ms \\
\midrule
\textbf{TIME PER OUTPUT TOKEN} & 155.95 ms & 73.81 ms & 18501.90 ms & 680.44 ms \\
\midrule
\textbf{TOKEN GENERATION TIME} & 10604.30 ms & 8192.60 ms & 1813186.29 ms & 70085.14 ms \\
\bottomrule
\end{tabular}
\\[0.05cm]
\footnotesize
\end{table*}

\section{Results and Discussions}

All experiments maintain consistent hardware configurations and inference parameters to ensure reproducible comparative analysis between resource constrained and larger models for accessibility focused video description generation.

\textbf{Table 1} reveals that SmolVLM2-500M demonstrates strong prompt sensitivity with clear performance patterns across indoor and outdoor scenarios. The Prompt + AD Guidelines approach dominates most evaluation metrics on both datasets, showing consistent alignment with AD-style references and superior lexical overlap performance. However,
 Prompt + Context + AD Guidelines occasionally excels in semantic-matching metrics like METEOR, indicating that contextual information can enhance meaning preservation. The model shows a notable bias toward AD-style instructions due to reference generation conditions and generally performs better on indoor Charades scenarios compared to outdoor AVCaps environments.

\textbf{Table 2} demonstrates that the larger 2.2B model exhibits different contextual utilization patterns. Table 1 presents results for all four prompting
strategies using the 500M model whereas Table 2 reports results for the two best-performing strategies:"Prompt + AD Guidelines" and "Prompt+Context+AD Guidelines".
 This decision was driven by Table 1's clear demonstration that
basic "Prompt Only" and "Prompt+Context" strategies consistently underperform compared to AD-enhanced approaches across all standard NLP metrics; therefore we omitted these two strategies for 2.2B model. In indoor scenarios, adding contextual information substantially enhances performance across all metrics, with Prompt + Context + AD Guidelines consistently outperforming the basic AD approach. This indicates the larger model can effectively exploit additional context to improve generation quality in structured, predictable environments. However, in outdoor scenarios, the performance gap narrows significantly, with context sometimes failing to provide meaningful improvements and occasionally diluting performance in precision-focused measures. 

\textbf{Table 3} examines how contextual integration affects description quality for BLV users. In indoor environments, adding context provides modest improvements in descriptiveness and clarity but introduces slight decreases in objectivity and more notable declines in accuracy, suggesting that enhanced vividness may come at the cost of strict factual reporting. Conversely, in outdoor environments, contextual cues prove particularly valuable, benefiting all evaluation dimensions with especially notable improvements in clarity and accuracy. This pattern indicates that contextual information helps BLV users gain a better awareness of space and is especially helpful in dynamic, visually complex outdoor environments.

\textbf{Table 4} reveals distinct strengths between model variants when using optimal prompting strategies. 
The 2.2B model demonstrates superior clarity and accuracy, along with better objectivity in indoor scenarios, making it more dependable for producing trustworthy, accessible descriptions, while the smaller model excels in descriptive richness. 

\textbf{Table 5} shows our llama-cpp inference framework, total latency comprises Load Time (model loading), Prompt Evaluation Time (input processing and tokenization), and Generation Time (step-by-step token generation). Latency depends on both per-token processing speed and the number of generated tokens. For the 500M INT8 model, quantization alters output probabilities due to reduced precision, leading to longer token sequences and increased Generation Time compared to FP32. Although the INT8 model achieves faster per-token processing (73.8 ms/token vs. 155.9 ms/token for FP32), it generates more tokens (111 vs. 68), resulting in higher overall latency.

\begin{table*}[!htbp]
\centering
\caption{\centering Quantitative Results for Multi-Context Evaluation Framework}
\label{tab:quant_results_context_eval}
\begin{tabular}{@{}llcccc@{}}
\toprule
\textbf{Dataset} & \textbf{Model Variant} & \textbf{Special Orientation} & \textbf{Social Interaction} & \textbf{Action \& Event} & \textbf{Ambience} \\
\midrule
Outdoor & 500M & 3.556 & 3.206 & 2.585 & 4.664 \\
 & 2.2B & 3.416 & 3.271 & 2.632 & 4.925 \\
Indoor & 500M & 3.223 & 3.281 & 2.126 & 4.318 \\
 & 2.2B & 2.976 & 3.332 & 1.949 & 3.532 \\
\bottomrule
\end{tabular}
\end{table*}
\begin{table*}[!htbp]
\centering
\caption{\centering Quantitative Results for Navigation Assistance Framework }
\label{tab:quant_results_asst}
\begin{tabular}{@{}llcccc@{}}
\toprule
\textbf{Dataset} & \textbf{Model Variant} & \textbf{Descriptiveness} & \textbf{Objective} & \textbf{Accurate} & \textbf{Clarity} \\
\midrule
Outdoor & 500M & 3.570 & 5.107 & 3.147 & 3.930 \\
 & 2.2B & 3.239 & 4.909 & 3.370 & 3.742 \\
Indoor & 500M & 3.258 & 5.02 & 3.002 & 3.533 \\
 & 2.2B & 2.808 & 5.193 & 3.519 & 3.478 \\
\bottomrule
\end{tabular}
\end{table*}
In \textbf{Table 6}, the multi-context evaluation framework shows that model size scaling does not uniformly improve performance across all contextual dimensions for BLV users. The 500M model demonstrates better performance in Ambience context description, showing that the smaller models are good at capturing environmental scenarios and visual mood essential for BLV spatial understanding.  The Action \& Event context consistently scores lowest across all model-dataset combinations, showing the critical limitation in temporal sequence description that affects BLV users' ability to follow dynamic content. 

\textbf{Table 7} demonstrate that the 500M model consistently outperforms the 2.2B variant in Objectivity scores, indicating that smaller models provide more factual, assumption-free descriptions crucial for BLV navigation safety. However, the larger model shows better accuracy performance in outdoor scenarios. 
The consistently moderate Descriptiveness scores  across two models reveal a critical gap in providing the detailed spatial information that BLV users require for effective navigation. 

Our NLP metric scores align with video captioning benchmarks: BLEU-1 > 0.3 indicates strong performance, and CIDEr > 0.7 is very good, 0.4–0.7 moderate, < 0.4 low (Vedantam et al., 2015). Our results show BLEU-1 ranging from 0.135 to 0.327 and CIDEr from 0.072 to 0.207, which fall within typical ranges for accessibility-focused video descriptions that demand richer contextual
detail. The 500M model demonstrates superior performance in outdoor scenarios and achieves higher objectivity scores (5.02-5.11) crucial for BLV safety, while the 2.2B model excels in indoor clarity (3.414 vs 3.094) and spatial accuracy. Action Events score lowest (1.95–2.63) due to VLMs’ difficulty with sequential temporal reasoning, while Descriptiveness (2.5–3.6) indicates limited spatial detail for safe BLV navigation.

\section{CONCLUSION}
Our comprehensive evaluation reveals three critical insights that challenge conventional assumptions about model scaling for accessibility applications.
We introduce two novel evaluation frameworks "Multi-Context BLV Framework" and "Navigational Assistance Framework" that systematically address the bias of reference-based metrics against BLV user preferences. These frameworks demonstrate that smaller models (500M parameters) often excel in environmental adaptability and objective description generation, while larger models (2.2B parameters) provide enhanced precision in structured scenarios. Mobile evaluation establishes the feasibility of edge deployment with 60-83 second inference times for 500M models on consumer hardware, addressing privacy and connectivity barriers that disproportionately affect BLV users.

The practical deployment of accessibility-focused VLMs on ubiquitous consumer technology represents a significant step toward democratizing video accessibility, providing BLV users with immediate, private, and contextually relevant video descriptions independent of internet connectivity or centralized services.

\bibliography{custom}
\bibliographystyle{acl_natbib}

\end{document}